# Leveraging Language Models for Automated Patient Record Linkage

Mohammad Beheshti[1,2], Lovedeep Gondara[3,4], Iris Zachary[1,2]

mbwnh@missouri.edu, Lovedeep.Gondara@phsa.ca, zacharyi@missouri.edu

1. Missouri Cancer Registry and Research Center, Department of Public Health, College of Health Sciences, University of Missouri, Columbia, MO, USA
2. MU Institute for Data Science and Informatics, University of Missouri, Columbia, MO, USA
3. Provincial Health Services Authority, Vancouver, Canada
4. School of Population and Public Health, University of British Columbia, Vancouver, Canada

## ABSTRACT

**Objective:** Healthcare data fragmentation presents a major challenge for linking patient data, necessitating robust record linkage to integrate patient records from diverse sources. This study investigates the feasibility of leveraging language models for automated patient record linkage, focusing on two key tasks: blocking and matching. **Materials and Methods:** We utilized real-world healthcare data from the Missouri Cancer Registry and Research Center, linking patient records from two independent sources using probabilistic linkage as a baseline. A transformer-based model, RoBERTa, was fine-tuned for blocking using sentence embeddings. For matching, several language models were experimented under fine-tuned and zero-shot settings, assessing their performance against ground truth labels. **Results:** The fine-tuned blocking model achieved a 92% reduction in the number of candidate pairs while maintaining near-perfect recall. In the matching task, fine-tuned Mistral-7B achieved the best performance with only 6 incorrect predictions. Among zero-shot models, Mistral-Small-24B performed best, with a total of 55 incorrect predictions. **Discussion:** Fine-tuned language models achieved strong performance in patient record blocking and matching with minimal errors. However, they remain less accurate and efficient than a hybrid rule-based and probabilistic approach for blocking. Additionally, reasoning models like DeepSeek-R1 are impractical for large-scale record linkage due to high computational costs. **Conclusion:** This study highlights the potential of language models for automating patient record linkage, offering improved efficiency by eliminating the manual efforts required to perform patient record linkage. Overall, language models offer a scalable solution that can enhance data integration, reduce manual effort, and support disease surveillance and research.

# INTRODUCTION

Despite significant advances, healthcare data remains fragmented, with patient records scattered across hospitals, laboratories, and electronic health systems. This fragmentation arises from the diverse range of biomedical data sources, including structured information like laboratory results and imaging, as well as unstructured data such as clinical notes and pathology reports (1). Central cancer registries serve as a prime example of healthcare organizations facing these challenges, as they aggregate data from multiple sources including hospitals, physician offices, and pathology labs to track cancer incidence, treatment, and outcomes (2).

To address this challenge, record linkage plays a critical role in integrating fragmented data, ensuring that all relevant medical information is associated with the correct individual. Record linkage or entity resolution is the process of connecting records from different data sources that refer to the same entity, such as a patient, to create a comprehensive view of that entity's information (3). Traditionally, record linkage is achieved through either deterministic or probabilistic methods. Deterministic record linkage relies on exact matches of specific identifiers, such as names or dates of birth, ensuring high precision but struggling with missing or inconsistent data. In contrast, probabilistic record linkage assigns weights to multiple identifiers, allowing for flexible matching despite variations, making it more effective in handling data inconsistencies (4,5). However, it requires careful definition of matching rules and threshold settings to determine what constitutes a reliable match, which can be challenging and may vary across datasets. Additionally, probabilistic methods often necessitate manual review to label uncertain matches, increasing the time and effort required for accurate record linkage (6).

Recent advancements in natural language processing (NLP) have introduced transformer-based language models capable of understanding complex text patterns (7,8), making them well-suited for automating entity resolution. Consequently, numerous studies have explored their potential to enhance and streamline this process. For example, Li et al. introduced Ditto, a system that fine-tunes transformer-based models like BERT and RoBERTa to enhance entity matching accuracy (9). Similarly, other researchers have leveraged transformer-based models with different approaches for product matching and general entity resolution tasks (10–14). Building on this progress, recent studies have explored large language models (LLMs) such as ChatGPT, Llama, and Mixtral for entity resolution, achieving impressive performance (10,11).

While language models have demonstrated success in general entity resolution using widely adopted benchmark datasets, their application to patient or individual record linkage remains underexplored. Notably, there has been no focused research on leveraging these models with real-world patient-level data, leaving a critical gap in understanding their effectiveness in this domain. This study aims to fill this gap by investigating the use of language models for patient record linkage using real-world healthcare data. By automating the record linkage process, this approach can reduce manual effort, save time and resources, and ultimately improve the availability and completeness of patient data for clinical, research, and public health registry operations. Automation can support more efficient disease surveillance, epidemiological studies, and health policy decision-making, ensuring that public health registries maintain comprehensive and high-quality data for monitoring population health trends and informing interventions.

This study aims for the following goals:

1. Assess the feasibility of using language models as a blocking technique in patient record linkage and compare their performance with the traditional methods of blocking.
2. Investigate the use of language models for matching patient records and evaluate their performance.
3. Identify key challenges in applying language models to patient record linkage.

## MATERIALS AND METHODS

### What is blocking?

Blocking is the first step in a record linkage project, playing a crucial role in enhancing efficiency by reducing the number of record comparisons in large datasets. When linking records across multiple databases without unique identifiers, a naive approach that compares all possible pairs would be computationally prohibitive, growing quadratically with the dataset size. Blocking mitigates this challenge by grouping records into smaller, more manageable subsets or "blocks" based on shared characteristics, allowing comparisons to be performed only within these blocks.

Traditional blocking methods rely on predefined attributes, such as exact matches on geographic location, birth year, or name components, to partition records into blocks. However, these approaches can be sensitive to data entry errors and variability in formatting. More advanced techniques, such as phonetic encoding (e.g., Soundex, Metaphone), clustering-based methods (e.g., Canopy clustering, Sorted Neighborhood), and similarity-based approaches (e.g., Jaro-Winkler, Levenshtein distance), improve robustness against errors by considering approximate similarities rather than exact matches (12). By implementing blocking effectively, organizations can significantly reduce the computational burden of record linkage while preserving accuracy, making it an indispensable step in data integration, de-duplication, and entity resolution tasks.

Advancements in natural language models and language models have introduced the use of embeddings to enhance blocking efficiency by capturing semantic similarity between records. Unlike traditional methods that rely on exact or approximate textual similarity, language model embeddings generate vector representations of textual data, allowing for more nuanced comparisons. Transformer-based models (e.g., BERT, RoBERTa, Sentence-BERT) can encode entity attributes into dense vector spaces where similar entities are positioned closer together. This enables more flexible and scalable blocking strategies, reducing reliance on manually crafted similarity metrics.

**What is matching?**

Matching is the process of assessing the similarity between candidate record pairs identified during blocking to determine whether they refer to the same entity. Traditional matching techniques fall into two main categories: deterministic and probabilistic. Deterministic matching relies on strict rule-based comparisons, considering records a match only if specific attributes, such as names or dates of birth, are identical. While this method is straightforward and precise, it is highly sensitive to inconsistencies like typos, formatting differences, or missing values (6)

Probabilistic matching provides a more flexible alternative by assigning weights to different fields and calculating a likelihood score for each potential match. This method is based on statistical models that estimate the probability of two records referring to the same entity by comparing field-level similarities. It considers both matching (agreement) and non-matching (disagreement) probabilities, typically using techniques such as Fellegi-Sunter's model (4), which applies Bayesian principles to compute match scores. Higher weights are given to fields that are more reliable for distinguishing records (e.g., Social Security numbers or full names), while fields prone to variation (e.g., address or phone number) are weighted lower. A final composite score determines whether a pair is classified as a match, a non-match, or a possible match requiring manual review.

To further enhance automation and minimize manual intervention, natural language processing (NLP)-based approaches are increasingly used in record matching. These methods frame matching as a binary classification task, where each candidate pair is evaluated and categorized as either a match or a non-match. By leveraging machine learning and deep learning techniques, NLP models can capture complex patterns in textual data, improving accuracy and scalability in entity resolution tasks.

**Data source and preprocessing**

This study utilized data from the Missouri Cancer Registry and Research Center. Two distinct data sources were selected for record linkage. Dataset A was extracted from the primary patient database of the cancer registry (CRS Plus), which contains consolidated records of patients diagnosed with cancer. Dataset B was obtained from the ePath Reporting database (eMaRC Plus), which includes pathology records submitted electronically by laboratories across the state of Missouri. Patient records from 2021 and 2022 were extracted from both databases. Data from 2022 was used to fine-tune the language models, while data from 2021 was utilized for testing the models.

To create the labeled linked data (denoted as AB), we used Match*Pro (13), a free software that applies the Fellegi and Sunter model (4) for probabilistic record linkage. The linkage configuration, detailed in Table 1, was utilized which was designed to maximize recall, providing a good training set as well as a baseline for our experiments.

Table 1. Blocking and Matching Strategies Used to Create Dataset AB via Probabilistic Linkage.

| Field Name | Blocking Strategy | Matching Strategy |
|---|---|---|
| First Name | Soundex (Phonetic Matching) | Jaro-Winkler Distance = 0.8 |
| Middle Name | - | Jaro-Winkler Distance = 0.8 |
| Last Name | Soundex (Phonetic Matching) | Jaro-Winkler Distance = 0.8 |
| Birth Date | Exact Match | Same Month and Year |
| Sex | - | Exact Match |
| Social Security Number | Exact Match | 2 Edits or Transpositions |

The linkage configuration was executed for both train and test datasets independently. The resulting record pairs were manually reviewed by experienced human annotators and labeled as Match or Non-Match based on their identifiers, including First Name, Middle Name, Last Name, Sex, Birth Date, Social Security Number (SSN), and Address. One of the key outputs of the probabilistic linkage process was the "Overall Similarity Score," which quantified the degree of agreement between record pairs, providing a comprehensive measure of their likelihood of being a match. We used this score as a label for record pair similarity as well as a proxy for measuring the difficulty level of classifying the candidate record pairs.

It is important to note that SSN and Address were missing in 97% and 81% of the records in dataset B (2022) respectively. As a result, the matching decisions primarily relied on the other available identifiers. Moreover, due to the absence of a unique patient identifier in dataset B, duplicate pairs were present in the resulting linkage. These duplicates were removed to prevent potential overfitting. A summary of the datasets and their record counts is presented in Table 2.

*Table 2. Summary of Record Counts in Each Dataset and Their Corresponding Labels.*

|              | Dataset A | Dataset B | Dataset AB | Labels |
|---|---|---|---|---|
| *Train (2022)* | 51,943 | 29,552 | 58,383 | Non-Match: 54,858<br>Match: 3,525 |
| *Test (2021)*  | 51,781 | 26,958 | 52,917 | Non-Match: 50,561<br>Match: 2,356 |

**Experiment 1: blocking model**

Blocking is the process of generating candidate record pairs. This is the first step in a record linkage pipeline, and it is crucial for minimizing the number of necessary comparisons. Embedding-based semantic similarity has been widely adopted in the literature as a blocking technique, leveraging dense vector representations to capture nuanced relationships between records (14–17). Given its effectiveness in reducing computational complexity while preserving recall, we opted to experiment with this approach for patient-level blocking to enhance efficiency.

We utilized the SentenceTransformers (SBERT) library (18) to fine-tune RoBERTa (19) for generating sentence embeddings, which were subsequently used to compute the semantic similarity between records. The primary objective of this step was to achieve high recall, ensuring that all potential matches were identified and retained for further comparison. To support this goal, fields with a high rate of missing data, including SSN and Address, were excluded from the blocking experiment, as their inclusion could have altered the embeddings of records with missing values in one dataset, potentially reducing the accuracy of the similarity computations. Any remaining missing values were replaced with an empty string to minimize their impact on the embeddings.

Each record pair in dataset AB was independently serialized into a string as follows:

Serialize (A) ::= [FirstName$_A$] [MiddleName$_A$] [LastName$_A$] [BirthDate$_A$] [Sex$_A$]

Serialize (B) ::= [FirstName$_B$] [MiddleName$_B$] [LastName$_B$] [BirthDate$_B$] [Sex$_B$]

Where A and B represent the attributes of records from dataset A and dataset B, respectively.

These serialized string pairs were fed into the model along with their corresponding labels. The labels were derived from the "Overall Similarity Score" obtained through the probabilistic linkage. The fine-tuning objective was to adjust the model so that embeddings of similar record pairs moved closer together, while non-similar pairs were pushed further apart in the embedding space. To achieve this, we used cosine similarity loss and applied mean pooling, which has been shown to be more effective than CLS pooling in capturing semantic similarities (20). The model was trained for 5 epochs with a batch size of 64, using the SBERT's default training arguments.

During evaluation, records in both individual test datasets (A and B) were serialized in the same manner as during the fine-tuning process. Then the fine-tuned model was utilized to generate embeddings for each serialized record using mean pooling. To identify similar records between the two datasets, embeddings from dataset A were queried against those from dataset B using the K-Nearest Neighbors (KNN) algorithm implemented via the FAISS library (21). FAISS is a scalable library optimized for fast nearest neighbor search in high-dimensional spaces, making it ideal for large-scale dense embeddings by accelerating similarity retrieval with efficient indexing structures. While the KNN algorithm retrieves a fixed number of nearest neighbors (K), this approach can inadvertently include records with low similarity scores. To address this, a cosine similarity threshold was incorporated into the pipeline. Only record pairs with cosine similarity scores exceeding the predefined threshold were retained, ensuring that less relevant pairs were filtered out. The resulting candidate record pairs were subsequently evaluated against the ground truth test set (dataset AB). Finally, the fine-tuned model was evaluated using different combinations of K and similarity thresholds to find the optimal combination.

**Experiment 2: matching model**

Matching is the second step in the record linkage pipeline. During this stage, candidate record pairs generated in the blocking step are compared and classified as either a match or a non-match. It is important to note that this experiment was conducted independently of Experiment 1, meaning that the candidate pairs from the blocking model were not used in the matching model. Instead, we utilized the labeled dataset AB to ensure that the matching model's results were not influenced by the blocking model. However, in an automated setting, blocking must be performed before classifying candidate pairs using the matching model.

We experimented with several open-source language models in both a zero-shot setting and a fine-tuned setting, as listed in Table 3. DeepSeek-R1-70B and Llama-3.3-70B were used exclusively for zero-shot inference to evaluate their generalization capabilities without extensive fine-tuning. While we also experimented with a few-shot approach, it consistently underperformed, as the models' decisions were overly influenced by the limited examples provided.

Table 3. List of Language Models Used in Experiment 1 Under Fine-Tuned and Zero-Shot Settings.

| Model Name | Fine-Tuned | Zero-Shot |
|---|---|---|
| RoBERTa-Base (19) | ✅ | ❌ |
| Llama-3.2-3B-Instruct (22) | ✅ | ❌ |
| Mistral-7b-instruct (23) | ✅ | ✅ |
| Llama-3.1-8B-Instruct (22) | ✅ | ✅ |
| Mistral-Small-24B-Instruct (23) | ✅ | ✅ |
| Llama-3.3-70B-Instruct (22) | ❌ | ✅ |
| DeepSeek-R1-Distill-Llama-70B (24) | ❌ | ✅ |

All the identifiers mentioned in the data source section were included in the matching model. Any missing identifiers, such as SSN or Address, were replaced with the word "Unknown" if absent from a record.

Since encoder-only classification models like RoBERTa require a different fine-tuning and inference setup than generative LLMs like Llama, each setup is explained comprehensively in the following paragraphs.

**Classification model setup**

We fine-tuned RobertaForSequenceClassification (25), which is based on roberta-base with an additional fully connected linear layer, known as the classification head. This layer maps the CLS token embeddings to the desired binary classification labels. A softmax activation function is applied to the output logits of the linear layer to produce probability scores for each class. Finally, the argmax function is used to select the class with the highest probability as the predicted label.

We adopted the serialization method used in Ditto (9) to structure the records for model input. Each individual record in dataset AB was serialized using the following format:

Serialize (A) ::= [COL] $attr_1$ [VAL] $val_1$ ... [COL] $attr_k$ [VAL] $val_k$

Serialize (B) ::= [COL] $attr_1$ [VAL] $val_1$ ... [COL] $attr_k$ [VAL] $val_k$

where [COL] and [VAL] are special tokens that denote the beginning of attribute names and their corresponding values, respectively. The serialized records were then concatenated as a pair using the following format:

Serialize (A, B) ::= [CLS] serialize(A) [SEP] serialize(B) [SEP]

where [SEP] is a special token used to separate the two serialized records, and [CLS] represents the classification token at the start of the sequence.

This model was fine-tuned using the training arguments listed in Table 1, except for the LoRA rank and alpha, which are not applicable to RoBERTa.

**Generative LLMs setup**

For fine-tuning and inference on the large language models, record pairs in dataset AB were formatted into a structured prompt using the template shown in Figure 1. At the time of fine-tuning, the correct response was included in the prompt as part of the 'Assistant' token.

```
prompt = f"""
You are given two patient records. Your task is to determine whether they
belong to the same individual. Consider factors such as name similarity, date
of birth, and other identifying attributes. Only respond with "Yes" or "No".

    Record 1:
        - First Name: {row['record1 First Name']}
        - Middle Name: {row['record1 Middle Name']}
        - Last Name: {row['record1 Last Name']}
        - Date of Birth: {row['record1 Date of Birth']}
        - SSN: {row['record1 SSN']}
        - Sex: {row['record1 Sex']}
        - Address: {row['record1 Address']}

    Record 2:
        - First Name: {row['record2 First Name']}
        - Middle Name: {row['record2 Middle Name']}
        - Last Name: {row['record2 Last Name']}
        - Date of Birth: {row['record2 Date of Birth']}
        - SSN: {row['record2 SSN']}
        - Sex: {row['record2 Sex']}
        - Address: {row['record2 Address']}

    """
```

*Figure 1. Prompt Template Used for Fine-Tuning and Inference on Generative Models*

All the generative LLMs were fine-tuned using the LoRA optimization technique (26) with the parameters specified in Table 4.

*Table 4. Training Arguments Used for Fine-Tuning the Models for the Matching Task*

| Parameter | Epochs | Batch Size | Learning Rate | Optimizer | LR Scheduler | Warmup Ratio | Weight Decay | LoRA Rank | LoRA Alpha |
|---|---|---|---|---|---|---|---|---|---|
| **Value** | 3 | 32 | $2e^{-5}$ | AdamW | Cosine | 0.1 | 0.01 | 32 | 32 |

The Unsloth library (27) was employed to enhance computational efficiency and optimize VRAM consumption, enabling scalable fine-tuning and inference of LLMs. The specific Unsloth-optimized versions of the LLMs used in this study are listed in Table 5.

Table 5. Unsloth-optimized versions of the LLMs used in this study

| Model Name | Hugging Face Link |
| --- | --- |
| Llama-3.2-3B-Instruct | https://huggingface.co/unsloth/Llama-3.2-3B-Instruct |
| Mistral-7b-instruct | https://huggingface.co/unsloth/mistral-7b-instruct-v0.3 |
| Llama-3.1-8B-Instruct | https://huggingface.co/unsloth/Meta-Llama-3.1-8B-Instruct |
| Mistral-Small-24B-Instruct | https://huggingface.co/unsloth/Mistral-Small-24B-Instruct-2501-unsloth-bnb-4bit |
| Llama-3.3-70B-Instruct | https://huggingface.co/unsloth/Llama-3.3-70B-Instruct-bnb-4bit |
| DeepSeek-R1-Distill-Llama-70B | https://huggingface.co/unsloth/DeepSeek-R1-Distill-Llama-70B-bnb-4bit |

All models except DeepSeek-R1 were used for inference with sampling disabled (temperature = 0) to enforce a deterministic approach. They were also provided with the system prompt, "You are a helpful assistant," during both fine-tuning and inference. For DeepSeek-R1, its developers recommended avoiding the system prompt and suggested using a temperature setting of 0.6 to prevent endless repetitions during the thinking process. The model outputs were then compared against the ground truth labels to assess performance.

All the experiments in this study were performed locally on a workstation with a single NVIDIA RTX A6000 (48GB VRAM) GPU, AMD Ryzen Threadripper PRO 5955WX (16 Cores) CPU, and 128 GB RAM.

Finally, the models were used for inference on the test record pairs (test dataset AB) and the answers were evaluated against the ground truth labels.

# RESULTS

**Record pairs distribution**

We utilized the "Overall Similarity Score" obtained from the probabilistic linkage to assess how matching and non-matching record pairs are distributed across different similarity levels. Figure 2 presents this distribution for both the train and test datasets.

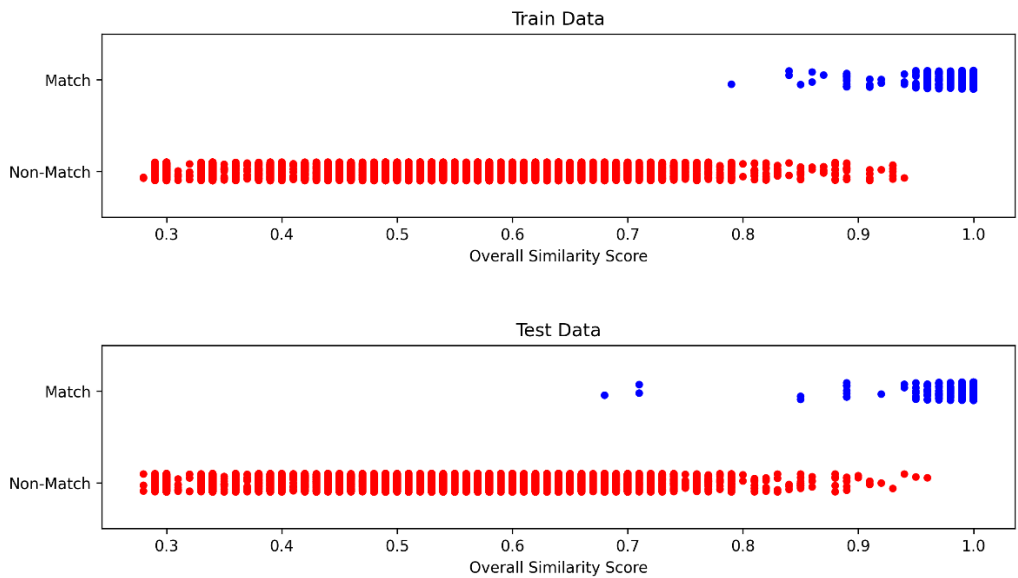

*Figure 2. Distribution of Matching and Non-Matching Records by Overall Similarity Score in Train and Test Data. Note: This strip plot contains a large number of data points, leading to overlapping dots that may cause certain regions to appear less dense than they are.*

Matching record pairs were highly concentrated at similarity scores above 0.8, with the majority of them clustering near 0.9 and higher. On the other hand, non-matching pairs, are widely dispersed across lower similarity ranges. An overlap between non-matches and matches occurs in the 0.85–0.95 similarity range, where false positives and false negatives are likely to appear. This ambiguity zone is where setting a fixed similarity threshold fails to achieve high accuracy, as some true matches may be excluded (false negatives), while certain non-matching pairs may be mistakenly classified as matches (false positives). Additionally, while the train and test datasets exhibit a similar distribution, applying a fixed threshold from one dataset to another may not yield optimal results. The presence of outliers in both matching and non-matching records can significantly impact classification performance, necessitating adaptive approaches.

**Blocking model performance**

The evaluation of the fine-tuned model for record blocking was conducted using different values of nearest neighbors (K) and cosine similarity thresholds to assess their impact on retrieval performance. The primary objective was to maximize recall while minimizing the number of candidate pairs generated. Figure 3 visualizes the variation in the number of candidate pairs generated and true matches missed as the cosine similarity threshold and k value change.

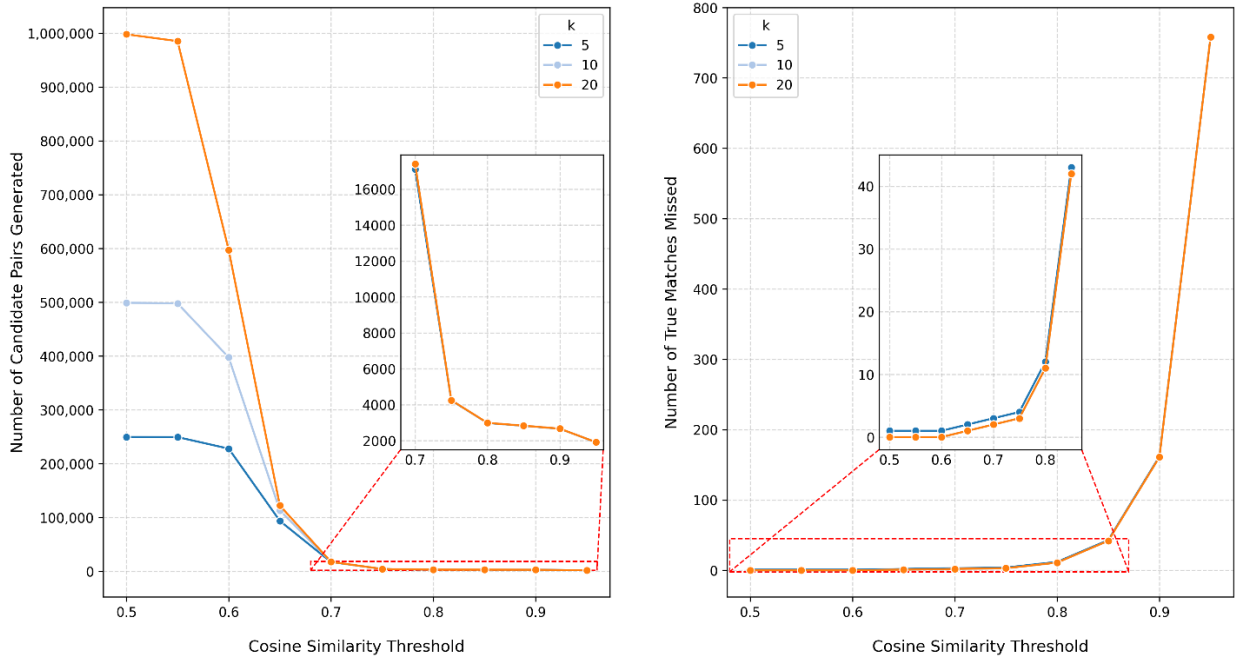

*Figure 3. Impact of Cosine Similarity Threshold and k on the Number of Candidate Pairs Generated and True Matches Missed.*

The results demonstrate that the similarity threshold is a critical factor in determining recall and retrieval efficiency. At lower similarity thresholds (0.5–0.6), recall remained near perfect (100%), ensuring nearly all true matches were retrieved, but at the cost of generating an excessive number of pairs, resulting in low efficiency. For instance, at a threshold of 0.5 with K=10, nearly 500,000 pairs were retrieved. Increasing the threshold to 0.7 significantly reduced the number of pairs generated to 17,396. However, this increase in efficiency came with a slight decline in recall, missing two true matches. As the threshold was raised further to 0.75, the number of retrieved pairs decreased significantly to 4,250, improving efficiency. Despite this substantial gain in efficiency, recall continued to stay almost the same, missing only three true matches. At higher similarity thresholds (0.8 and above), the trade-off between recall and efficiency became more pronounced. The number of retrieved pairs dropped sharply, improving efficiency. However, this came at the cost of a significant recall reduction, declining from 99.52% at 0.8 to just 67.04% at 0.95, with 758 true matches missing. These results demonstrate that while higher thresholds improve efficiency, excessive filtering leads to substantial recall loss. Our results indicate that the number of nearest neighbors (K) also influenced retrieval performance. Increasing K from 5 to 30 significantly increased the number of candidate pairs generated, particularly at the 0.6 threshold and below. However, this effect became negligible beyond the 0.7 threshold. Additionally, increasing K from 5 to 10 slightly improved recall, ensuring full retrieval, but further increases beyond 10 had no additional impact on recall.

Overall, the results indicate that, in our case, an optimal balance between recall and efficiency can be achieved using K=10 and a cosine similarity threshold of 0.75, which maintains high recall while significantly reducing the number of retrieved candidate pairs, ensuring a more efficient matching process. Conducting similar exploratory analyses is essential for anyone seeking to determine optimal thresholds for their specific use case. As discussed in the methods section, the resulting candidate record pairs were not used in the matching experiment.

**Matching model performance**

In this experiment, we assessed different language models for record matching by comparing their performance metrics using the input datasets described in the data source and preprocessing section. Table 6 presents the performance of the models under both fine-tuned and zero-shot settings. Since all models achieved nearly perfect F1 scores, this metric was not particularly useful for distinguishing their performance. Therefore, we focused on the total number of false positives and false negatives (FP + FN) as a more informative measure of how many total incorrect predictions each model made.

Table 6. Performance of Matching Models Under Fine-Tuned and Zero-Shot Settings.

| Model | Fine-Tuned | | | | Zero-Shot | | | |
|---|---|---|---|---|---|---|---|---|
| | FP | FN | FP + FN | F1 Score | FP | FN | FP + FN | F1 Score |
| RoBERTa-Base | 8 | 19 | 27 | 0.996995 | — | — | — | — |
| Llama-3.2-3B | 1 | 8 | 9 | 0.998999 | — | — | — | — |
| Mistral-7B | 0 | 6 | **6** | **0.999333** | 2,450 | 4 | 2,454 | 0.816149 |
| Llama-3.1-8B | 3 | 5 | 8 | 0.999111 | 0 | 839 | 839 | 0.887572 |
| Mistral-Small-24B | 0 | 11 | 11 | 0.998776 | 0 | 55 | **55** | **0.993823** |
| Llama-3.3-70B | — | — | — | — | 14 | 65 | 79 | 0.991135 |

Our results show that among the fine-tuned models, Mistral-7B had the lowest number of incorrect predictions (FP + FN = 6), followed by Llama-3.1-8B (FP + FN = 8), Llama-3.2-3B (FP + FN = 9), and Mistral-Small-24B (FP + FN = 11). RoBERTa had the highest error count among the fine-tuned models (FP + FN = 27).

In the zero-shot setting, Mistral-Small-24B achieved the best performance, with the lowest number of incorrect predictions (FP + FN = 55) and an F1 score of 0.994, indicating strong generalization without fine-tuning. Llama-3.3-70B followed closely, with 79 incorrect predictions and an F1 score of 0.991. Llama-3.1-8B and Mistral-7B ranked third and fourth respectively, showing a notable decline in F1 score compared to the better-performing, larger models.

DeepSeek-R1-70B is a reasoning model that excels in complex tasks, following chain-of-thought process, where it generates thinking tokens before outputting the final answer. This makes its inference computationally expensive, time consuming and impractical for a task involving tens of thousands of records. Hence, we only evaluated this model on a subset of the test data (N = 2736) where [0.65 < Overall Similarity Score < 1.0]. This threshold was chosen to exclude record pairs that were not challenging to classify, thereby significantly reducing the required computational resources and time by keeping only the edge cases. DeepSeek-R1-70B achieved an F1 score of 0.855, with 66 false positives and 32 false negatives (FP + FN = 98). In contrast, on the same subset, Mistral-Small-24B demonstrated superior performance, achieving an F1 score of 0.93, with zero false positives and 38 false negatives (FP + FN = 38). Llama-3.3-70B followed closely, attaining an F1 score of 0.909 with 7 false positives and 43 false negatives (FP + FN = 50).

It is also important to highlight the significant difference in processing time between the two models. DeepSeek-R1-70B took approximately 26 hours to process the subset, whereas Llama-3.3-70B completed the same subset in 30 minutes.

## DISCUSSION

Our study evaluated the feasibility of leveraging language models for automated patient record linkage, focusing on two critical steps: blocking and matching. Each step is discussed separately in the following paragraphs.

### Language models for blocking

The use of fine-tuned RoBERTa embeddings for record blocking demonstrated good performance in reducing the number of candidate pairs generated while maintaining a near perfect recall. For example, using a cosine similarity threshold of 0.75 with a K=10, the blocking model generated 4,250 candidate pairs. This is a 92% reduction compared to the baseline test dataset with 52,917 candidate pairs. While the blocking model achieved significant improvement in efficiency compared to the traditional rule-based blocking, it sacrificed a few true matches. This might not be ideal depending on the use case of the linkage. For example, at central cancer registries, record linkage is used to identify unreported cancer cases by comparing the existing patient records to the records submitted by the pathology laboratories. Therefore, achieving a 100% blocking recall is essential to avoid missing any cancer cases. However, if the linkage is used for research rather than for disease surveillance, a slight loss in recall may be acceptable.

Despite significantly reducing the number of candidate pairs, the blocking model showed limitations in handling minor typos. Upon examining the cosine similarity scores of record pairs, we observed that the blocking model sometimes assigns low similarity scores to nearly identical records that contain only minor typos. This likely happens because most language model tokenizers, including RoBERTa, operate at the subword level, breaking words into smaller subword units rather than individual characters or full words. As a result, even a minor single-character typo can cause the tokenizer to split an identifier, such as the name, into two unrelated subwords, shifting the entire record to a distant position within the embedding space and reducing its cosine similarity score. This suggests that embedding-based semantic similarity approach may be suboptimal for patient record blocking. Future research should explore methodologies that incorporate character-level tokenizers to enhance robustness against minor typos.

Given these challenges, an alternative approach that combines traditional rule-based blocking with probabilistic linkage may offer a more effective solution. Our findings indicate that when traditional rule-based blocking is combined with the "Overall Similarity Score" threshold obtained from probabilistic linkage, the rule-based approach can outperform the language model-based approach. For instance, in the test dataset shown in Figure 2, no matches were found below the 0.65 threshold, and no non-matches had an Overall Similarity Score of 1.0. This range could be leveraged as a confidence threshold to filter out easily classifiable pairs. As a result, the dataset would be reduced to 2,736 candidate pairs, representing a 95 percent reduction compared to the baseline test dataset of 52,917 candidate pairs, while still maintaining 100% blocking recall.

Overall, while language models provide a flexible approach to patient record blocking without the need to define meticulous blocking strategies, they do not achieve the highest accuracy or computational efficiency compared to a hybrid rule-based and probabilistic approach.

**Language models for matching**

All the fine-tuned models demonstrated strong performance on unseen data, with Mistral-7B achieving the fewest incorrect predictions (FP + FN = 6), followed closely by Llama-3.1-8B (FP + FN = 8) and Llama-3.2-3B (FP + FN = 9). While RoBERTa also performed well, its higher error rate suggests that the more advanced architectures of LLMs may be better suited for understanding the nuances of this task. Notably, generative LLMs such as Mistral and Llama were not specifically designed for classification, unlike conventional encoder-only models like RoBERTa, which include a dedicated classification head. This finding highlights the potential of instruction fine-tuning LLMs in achieving superior results for record linkage compared to traditional classification models.

Our results show that fine-tuned language models consistently outperform their zero-shot counterparts, underscoring the importance of domain-specific fine-tuning for specialized tasks like patient record linkage. Notably, a fine-tuned smaller model, such as Mistral-7B or Llama-3.2-3B, can even surpass larger non-fine-tuned models, demonstrating that fine-tuning enhances efficiency and effectiveness in targeted applications. Moreover, Mistral-Small-24B achieved the best zero-shot performance despite being smaller than Llama-3.3-70B. This suggests that factors such as training data, model architecture, and optimization can sometimes outweigh sheer parameter count. Nonetheless, both larger models still outperformed smaller ones in the zero-shot setting, aligning with the fact that increasing model size generally enhances zero-shot performance and improves generalization across diverse tasks (28–31).

Our evaluation of DeepSeek-R1-70B, conducted exclusively in the zero-shot setting on a smaller yet more challenging subset of the test data, revealed significant differences in performance and computational cost compared to its closest non-reasoning counterpart, Llama-3.3-70B. While DeepSeek-R1-70B achieved a respectable F1 score of 0.855, its total error count (FP + FN = 98) was double that of Mistral-Small-24B (FP + FN = 38) and Llama-3.3-70B (FP + FN = 50). The superior performance of the non-reasoning models highlights their greater effectiveness in the patient record linkage task. This finding is particularly intriguing, given that reasoning models like DeepSeek-R1-70B have demonstrated strong performance in tasks requiring logical inference and complex reasoning (32). Beyond predictive performance, we observed a significant difference in computational efficiency between reasoning and non-reasoning models. DeepSeek-R1-70B required approximately 26 hours to process the subset, whereas the non-reasoning model, Llama-3.3-70B, processed the same subset in 30 minutes. Both models were used for inference in batches; however, non-reasoning models accommodate larger batch sizes because they do not generate thinking tokens and require a smaller context window. This allows them to process the record pairs more efficiently. Additionally, reasoning models take longer to generate their final decision due to the step-by-step reasoning process, while non-reasoning models could be enforced to generate a single "Yes" or "No" token, further contributing to their faster processing time. This significant difference in the inference time highlights the computational demands of reasoning models like DeepSeek-R1 and suggests that, despite their potential, their practicality for large-scale record linkage tasks may be limited due to their extended inference time. These findings underscore the importance of balancing predictive accuracy with computational efficiency when selecting language models for real-world applications.

## CONCLUSION

This study demonstrates the potential of language models to enhance patient record linkage by automating both blocking and matching tasks. While the embedding-based blocking model showed good performance, it struggled with minor data inconsistencies, indicating that alternative approaches, such as hybrid probabilistic blocking, may be better suited for structured patient records. Fine-tuned generative LLMs outperformed traditional and zero-shot models in matching accuracy, offering a robust solution with minimal errors. However, the trade-off between predictive performance and computational efficiency must be considered when deploying these models at scale. Overall, large language models show strong potential for automating the matching of patient records, offering improved efficiency by eliminating the manual efforts required for record linkage. This approach can enhance the integration and availability of healthcare data, supporting disease surveillance and research.

## AUTHOR CONTRIBUTIONS

Mohammad Beheshti (Conceptualization, Study design, Implementation, Writing—original draft), Lovedeep Gondara (Study design, Implementation, Writing—review and editing), Iris Zachary (Study design, Writing—review and editing, Supervision, Resources).

## FUNDING

This work was supported by the Missouri Cancer Registry and Research Center under contract number NU58DP007130-03, as well as a surveillance contract between the Missouri Department of Health and Senior Services (DHSS) and the University of Missouri.

## CONFLICT OF INTEREST

The authors have no competing interests to declare.

## DATA AVAILABILITY

The data used in this study contains Protected Health Information (PHI) and cannot be shared due to privacy regulations and ethical considerations.